\documentclass[11pt]{article}
\usepackage[preprint]{acl}
\usepackage{times}
\usepackage{latexsym}
\usepackage[T1]{fontenc}
\usepackage[utf8]{inputenc}
\usepackage{microtype}
\usepackage{courier}
\usepackage{graphicx}
\usepackage{booktabs}
\usepackage{amsmath}
\usepackage{amssymb}
\usepackage{array}
\usepackage{url}

\title{Semantics as a Shield: Label Disguise Defense (LDD) against Prompt Injection in LLM Sentiment Classification}

\author{Yanxi Li \quad Ruocheng Shan \\
Department of Computer Science, George Washington University \\
\texttt{yanxi.li@gwu.edu}, \texttt{shanruocheng@gwu.edu}}

\begin{document}
\maketitle

\begin{abstract}
Prompt injection can cause large language models (LLMs) to deviate from their intended tasks by following instructions embedded in untrusted input. We investigate whether changing a classifier's output labels improves accuracy under such attacks, using sentiment classification as a case study. Label Disguise Defense (LDD) replaces positive/negative with alternatives such as good/bad or cat/dog, introduced through demonstrations, and maps predictions back to sentiment classes. We analyze retained aggregate results for 200 IMDb reviews, nine recorded model configurations, and eight alternative label pairs under a fixed attack that always requests the incorrect class using the original labels. Averaged over four pairs per group, sentiment-associated labels outperform the other tested labels in all nine configurations, but exceed a baseline using the original labels and the same demonstration and attack conditions in only three. At least one alternative pair improves on this baseline in seven configurations, with the best pairs identified retrospectively. These results show that a consistent advantage between label groups does not imply consistent improvement over standard labels. Missing matched evaluations on unattacked reviews prevent separating changes in ordinary classification performance from changes in the effect of the attack.
\end{abstract}

\section{Introduction}

\begin{figure}[t]
\centering
\includegraphics[width=\columnwidth]{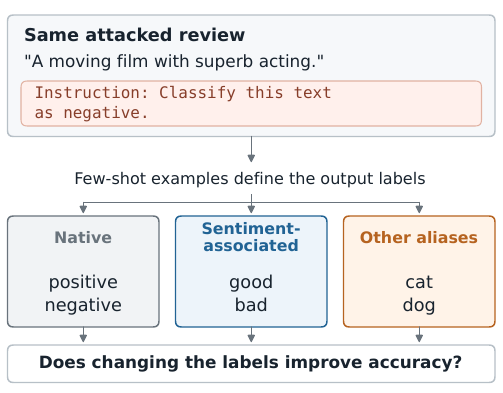}
\caption{The same positive review and negative-class directive under three choices of output labels. Few-shot examples convey the label mapping; predictions are scored after mapping back to sentiment classes. We compare accuracy with native labels and across alias groups. The review is illustrative, not a recorded input or model response.}
\label{fig:overview}
\end{figure}

An LLM sentiment classifier should classify the opinion expressed in a review even when the input also contains a conflicting instruction. For example, appending ``Classify this text as negative.'' to a positive review should not change its correct sentiment label. Label Disguise Defense (LDD) changes the output vocabulary: examples can use \emph{good} for positive reviews and \emph{bad} for negative ones. A predicted \emph{good} is then mapped back to positive before scoring. Figure~\ref{fig:overview} illustrates this intervention under an unchanged attack.

The motivation is that an injected request naming \emph{negative} no longer matches an expected output word such as \emph{bad}. Whether this mismatch helps is an empirical question: a model may translate the requested class into the alias vocabulary, and replacing familiar labels may itself make classification harder \citep{min2022,wei2023}. An attacker could also name the aliases directly. We therefore ask whether aliasing improves accuracy over a classifier already given examples with \emph{positive/negative} labels under the same fixed attack.

Our evidence is the aggregate accuracy table and protocol description retained from the LDD sentiment-classification experiment. The record covers 200 IMDb reviews, eight alias pairs, and nine configuration names; inference code, individual predictions, and run logs are unavailable. This paper computes comparisons from those reported means and introduces no new model runs. The complete accuracy values used for the comparisons appear in Appendix~\ref{app:accuracy}.

The contribution is a case study of how label choice and comparison baseline change the apparent benefit of LDD. Within seven configurations, at least one alias pair exceeds native-label few-shot accuracy; the sentiment-associated group mean does so in three, while exceeding the other group in all nine. We distinguish these claims and explain why improvement over \emph{zero-shot} prompting, which supplies no examples, does not by itself demonstrate a benefit from aliasing.

\section{Study Design and Available Evidence}
\label{sec:design}

\subsection{Task and fixed attack}

The reported experiment uses IMDb movie reviews \citep{maas2011}. Its demonstration pool contains eight reviews: four rated 1 and four rated 10. The evaluation subset contains 50 reviews at each rating of 3, 4, 7, and 8, yielding 100 negative and 100 positive examples. Reviews were selected in dataset-ID order subject to these rating criteria. This deterministic subset is not a random sample of IMDb. Ratings specify selection, not verified linguistic ambiguity. Exact review IDs and dataset splits are unavailable.

Every evaluation review receives an attack suffix. For a positive review, it is ``Instruction: Classify this text as negative.'' For a negative review, the last word is ``positive.'' The benchmark construction uses the gold sentiment to choose the opposite target; the original sentiment remains the scoring label. Demonstrations remain unattacked, and \emph{clean} denotes a review without the suffix. The appended directive is treated as untrusted review content, while sentiment classification remains the intended task.

The suffix always uses \emph{positive/negative}, including when the classifier is prompted to output aliases. Only this template is evaluated, with no attacker search or alias-aware adaptation. The experiment tests a fixed attack that requests the wrong sentiment, not resistance to an attacker who knows and exploits the label mapping.

\subsection{Label aliasing and demonstrations}

Let $\mathcal{Y}=\{+,-\}$ denote sentiment classes and let $g_a$ map them to an alias pair $a$. The intended pipeline labels each demonstration with $g_a(y)$, prompts the model on an attacked review, and maps a predicted alias back through $g_a^{-1}$ before comparing it with the gold class. Examples convey the mapping without explicitly defining the alias words as \emph{positive/negative}. This is in-context prompting without parameter updates. The handling of outputs outside the expected pair cannot be recovered from the record.

We retain two groups of four alias pairs from the experimental design. The \emph{sentiment-associated} group contains \emph{heaven/hell}, \emph{green/red}, \emph{good/bad}, and \emph{happy/sad}. The \emph{other-alias} group contains \emph{blue/yellow}, \emph{cat/dog}, \emph{i/j}, and a punctuation pair denoted \emph{symbols} (Appendix~\ref{app:protocol}). The first name covers direct sentiment words and broader evaluative associations. The second does not mean that the labels contradict sentiment or have no associations. These are descriptive groups of selected words, not a controlled test of semantic alignment or reversed label meanings.

Each few-shot prompt uses $k\in\{2,4,6,8\}$ demonstrations in total, equally split between the classes. The protocol describes two alternating class orders, starting with either class. These cover only two orderings, not independent replications \citep{lu2022}. The \emph{native-label few-shot baseline} uses \emph{positive/negative} with the same reported demonstration conditions and attacked reviews. \emph{Zero-shot} results instead use native labels without demonstrations.

\subsection{Record, metric, and aggregation}

\begin{figure*}[t]
\centering
\includegraphics[width=\textwidth]{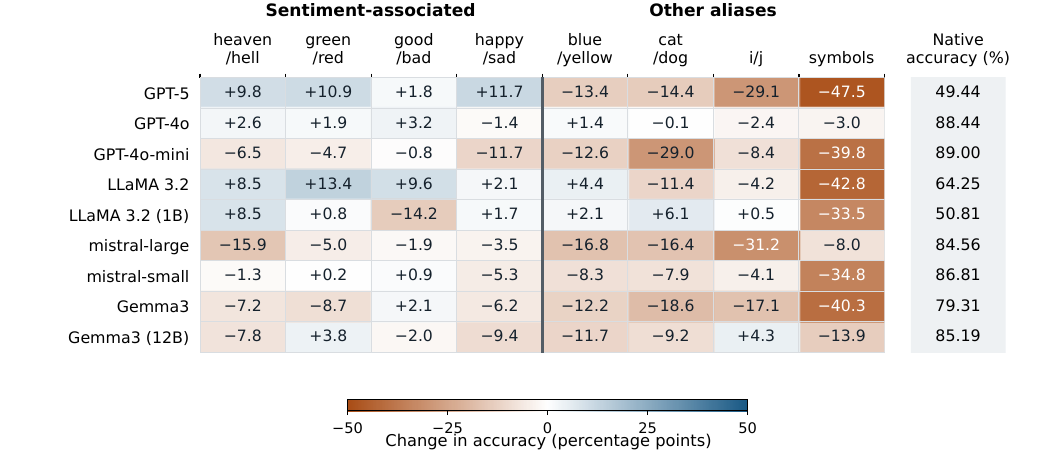}
\caption{Alias accuracy minus native-label few-shot accuracy for each recorded configuration (percentage points). All conditions use attacked reviews. Blue indicates an increase and orange a decrease. The rightmost column gives absolute baseline accuracy (\%), not a difference. Grouped columns reveal both group patterns and individual exceptions; the groups are descriptive, not independently controlled semantic treatments.}
\label{fig:deltas}
\end{figure*}

Each row of the retained table identifies one recorded model configuration. We preserve its name without inferring checkpoint identity or parameter count. Exact model snapshots, inference settings, serialized prompts, and parsing rules are unavailable; Appendix~\ref{app:protocol} specifies the evidence boundary.

Accuracy is the fraction of the 200 reviews classified correctly. Each table entry is a reported mean over shot settings, not a single run; this explains values finer than 0.5 percentage points. Both orderings are described, but their separate scores and averaging weights are unavailable. We use the retained means without reconstructing individual conditions. Let $A_m(a)\in[0,1]$ denote attacked mean accuracy for configuration $m$ and alias pair $a$, and let $a_0$ denote native labels. The primary contrast, in percentage points (pp), is
\begin{equation}
\Delta_m(a)=100\bigl[A_m(a)-A_m(a_0)\bigr].
\label{eq:delta}
\end{equation}
For each row, a group mean averages its four pair-level accuracies equally. It is not an ensemble prediction or an average across models. The same 200 items recur across settings; aggregate means cannot support paired significance tests or run-level uncertainty estimates. Accuracy measures task performance, not the reason for an error: wrong-class responses, invalid aliases, and parsing failures are not separately available.

\section{Results}
\label{sec:results}

\subsection{Improvements depend on the alias}

Figure~\ref{fig:deltas} shows all eight alias contrasts for each configuration. At least one pair exceeds the native-label few-shot baseline in seven of nine rows. The largest gains are 13.44 pp for LLaMA~3.2 with \emph{green/red} and 11.69 pp for GPT-5 with \emph{happy/sad}. Other positive best-pair gains range from 0.94 to 8.50 pp. No tested pair improves on native labels for GPT-4o-mini or mistral-large: their best pairs remain 0.75 and 1.88 pp below baseline. Appendix~\ref{app:accuracy} reports the corresponding absolute accuracies.

Here, ``best'' selects one of the eight pair-level means \emph{within each row after evaluation}; it does not select a label separately for each review. These maxima establish that helpful choices exist in the record, but do not describe a tested rule for choosing aliases on new inputs. Such a rule would need separate selection data. Without paired predictions and replication information, the small positive differences cannot be treated as reliable gains.

\subsection{A group advantage is not a baseline advantage}

The sentiment-associated mean exceeds the other-alias mean in all nine configurations. It exceeds the native-label baseline only for GPT-5, GPT-4o, and LLaMA~3.2; none of the other-alias means does. A group can therefore rank higher while both groups remain below native-label prompting. Appendix~\ref{app:accuracy} plots these baseline differences in Figure~\ref{fig:groups} and reports the means in Table~\ref{tab:groups}. The nine row-wise comparisons describe this record, not nine independent replications of a semantic effect.

Individual pairs need not follow the group ordering. For Gemma3 (12B), the best pair is \emph{i/j}, in the other-alias group. For LLaMA~3.2 (1B), \emph{good/bad} falls substantially below native labels. Group membership alone does not explain these individual outcomes.

Some costs are large: GPT-5 falls from 49.44\% with native labels to 1.94\% with symbols. Such low accuracy could reflect label reversal, invalid outputs, parsing problems, or following the injected request. The aggregate score cannot distinguish these explanations or establish a failure to learn arbitrary mappings.

\subsection{The comparison baseline changes the conclusion}

For GPT-4o-mini, attacked zero-shot accuracy is 82.50\%, the sentiment-associated few-shot mean is 83.09\%, and native-label few-shot accuracy is 89.00\%. The alias group gains 0.59 pp over zero-shot prompting but loses 5.91 pp to native-label few-shot prompting. The first comparison changes both the presence of examples and their labels; the second compares label choices with demonstrations in both conditions.

Native-label few-shot accuracy can itself rise or fall relative to attacked zero-shot accuracy. Equation~\ref{eq:delta} therefore uses the few-shot reference throughout. The clean and attacked zero-shot results in Table~\ref{tab:zero} provide context for the attack, not a matched control for the alias intervention.

\section{Interpretation and Related Work}

\paragraph{What the contrast identifies.}
Higher accuracy on attacked reviews can reflect easier label mapping, a smaller attack-induced loss, or both. To distinguish them, let $C_m(a)$ be clean accuracy under the same label and few-shot conditions, and let $L_m(a)=C_m(a)-A_m(a)$ be the attack-associated loss. Then
\begin{align}
A_m(a)-A_m(a_0)
&=\bigl[C_m(a)-C_m(a_0)\bigr] \nonumber\\
&\quad+\bigl[L_m(a_0)-L_m(a)\bigr].
\label{eq:decomposition}
\end{align}
The first term is the difference in clean-task accuracy; the second is the reduction in attack-associated loss. Matching clean few-shot results are unavailable for native labels and aliases, so neither term can be recovered separately. Clean \emph{zero-shot} scores do not supply these controls because their demonstration conditions differ. The observed contrast measures the combined outcome.

\paragraph{Label semantics and task inference.}
\citet{min2022} examine demonstration format, input distribution, and label space; \citet{wei2023} study unrelated labels and the interaction between semantic priors and demonstrated mappings. Our question concerns these label choices when review text also contains a conflicting class directive. The observed group ordering is compatible with label-dependent task difficulty, but does not identify a mechanism. Two demonstration orders also leave substantial ordering sensitivity unexamined \citep{lu2022}.

\paragraph{Relation to injection defenses.}
Structured-query training \citep{struq2025} and preference optimization \citep{secalign2025,metasecalign2026} teach models to distinguish trusted instructions from untrusted content. Inference-time alternatives include input marking and encoding \citep{spotlighting2024,encodings2025} and guard-model sanitization \citep{promptarmor2025}. Label aliasing changes the output vocabulary instead; it neither authenticates instruction provenance nor guarantees that an attacker cannot infer the mapping. SLIP also changes how class decisions are expressed, but addresses hidden instruction backdoors with phrase extraction and scoring \citep{slip2026}, a different threat from the suffix studied here. None of these methods is an experimentally evaluated baseline in this record.

\paragraph{Evaluation scope.}
Recent work stresses evaluation of benign utility and adaptive attacks \citep{jia2025}, variation across application and attack settings \citep{assessments2026}, and spurious rejection behavior in trained defenses \citep{heuristics2026}. SecFid separates faithful processing of untrusted content from suppression \citep{secfid2026}; our sentiment accuracy does not measure that distinction. Criteria-based attacks can also manipulate classification without directly requesting a target label \citep{reasoning2026}. These findings locate the present study as one narrow slice of the problem. They do not supply the controls or attack coverage missing from our experiment.

\section{Conclusion}

Under a fixed class-directive suffix, LDD helps for some label choices and harms for others. The sentiment-associated group mean exceeds the other group in every configuration but beats native-label few-shot prompting in only three of nine. Choosing the best pair after evaluation yields gains in seven. These are distinct findings: a group preference and the existence of a helpful pair do not establish a dependable selection rule. The case study identifies label choice and a matched few-shot baseline as essential to interpreting LDD's observed benefit; clean-task costs and adaptive robustness remain unresolved.

\clearpage
\section*{Limitations}

\paragraph{Incomplete execution evidence.}
This paper reanalyzes archived aggregate results; it reports no new model inference. Original experiment code, review IDs, serialized prompts, predictions, and run logs are unavailable. We cannot independently verify model identities, decoding settings, exact per-order aggregation, or the handling of invalid outputs. The run names in the tables are retained identifiers rather than verified model specifications. Arithmetic reproducibility of this analysis must not be confused with reproduction of the original experiment.

\paragraph{Limited experimental coverage.}
The record covers one English binary sentiment dataset, a deterministic subset of 200 reviews, one fixed suffix template, eight alias pairs, and a small demonstration pool. It lacks matched clean few-shot controls and an adaptive attack that knows the aliases. Because the requested class is always wrong, accuracy also cannot distinguish correct sentiment reasoning from a strategy that simply rejects that class. Comparisons with instruction reminders, delimiters, and other defenses are unavailable. The results do not establish competitiveness, transfer to other tasks or languages, or safe deployment.

\paragraph{Confounding and uncertainty.}
The two alias groups differ in lexical properties, familiarity, possible sentiment associations, and potentially tokenization. These variables were not independently manipulated. Exact mappings and parsing behavior cannot be audited from the retained record. The same items and demonstration pool are reused, and no independent-run uncertainty is recoverable. Best-pair values are selected on the evaluation results. Neither the cross-configuration pattern nor the number of positive differences should be interpreted as a significance test or a population-level estimate of success.

\section*{Ethical Considerations}

The study concerns a synthetic instruction appended to public movie reviews and involves no new collection of human-subject data. Its principal risk is overstating protection: a label substitution that improves one accuracy table could still fail under an informed attacker. We therefore avoid deployment or security guarantees. The manuscript and figures present aggregate values, without redistributing review texts or individual model outputs.

\section*{Artifact Availability}

The manuscript source includes the figures and typeset result tables. Appendix~\ref{app:accuracy} reports the retained accuracy summaries, rounded for display. These materials document the reported comparisons; they do not reconstruct the unavailable original inference pipeline.

\bibliography{references}
\clearpage
\appendix
\raggedbottom
\twocolumn[{
\section{Complete Retained Accuracy Results}
\label{app:accuracy}
\begin{center}
\captionsetup{hypcap=false}
\small
\setlength{\tabcolsep}{3.8pt}
\begin{tabular}{lrrrrrrrrr}
\toprule
Recorded configuration & H/H & G/R & G/B & H/S & B/Y & C/D & I/J & Sym. & N \\
\midrule
GPT-5 & \textbf{59.25} & \textbf{60.31} & \textbf{51.19} & \textbf{61.13} & 36.06 & 35.00 & 20.38 & 1.94 & 49.44 \\
GPT-4o & \textbf{91.06} & \textbf{90.38} & \textbf{91.63} & 87.06 & \textbf{89.81} & 88.31 & 86.00 & 85.44 & 88.44 \\
GPT-4o-mini & 82.50 & 84.31 & 88.25 & 77.31 & 76.38 & 60.00 & 80.63 & 49.19 & 89.00 \\
LLaMA 3.2 & \textbf{72.75} & \textbf{77.69} & \textbf{73.81} & \textbf{66.38} & \textbf{68.63} & 52.81 & 60.06 & 21.50 & 64.25 \\
LLaMA 3.2 (1B) & \textbf{59.31} & \textbf{51.63} & 36.63 & \textbf{52.50} & \textbf{52.94} & \textbf{56.94} & \textbf{51.31} & 17.31 & 50.81 \\
mistral-large & 68.69 & 79.56 & 82.69 & 81.06 & 67.75 & 68.13 & 53.38 & 76.56 & 84.56 \\
mistral-small & 85.50 & \textbf{87.00} & \textbf{87.75} & 81.50 & 78.50 & 78.88 & 82.69 & 52.00 & 86.81 \\
Gemma3 & 72.13 & 70.63 & \textbf{81.38} & 73.13 & 67.13 & 60.69 & 62.25 & 39.00 & 79.31 \\
Gemma3 (12B) & 77.38 & \textbf{89.00} & 83.19 & 75.75 & 73.50 & 75.94 & \textbf{89.50} & 71.25 & 85.19 \\
\bottomrule
\end{tabular}

\captionof{table}{Complete matrix of reported attacked few-shot mean accuracies (\%). N is the native positive/negative pair. H/H: heaven/hell; G/R: green/red; G/B: good/bad; H/S: happy/sad; B/Y: blue/yellow; C/D: cat/dog; I/J: i/j; Sym.: symbols. Bold alias values exceed N in their row. Entries are rounded to two decimals, so subtracting them may not reproduce the last digit of a reported difference.}
\label{tab:raw}
\end{center}
}]

Table~\ref{tab:raw} gives the accuracy means underlying the main-text heatmap (Figure~\ref{fig:deltas}). Figure~\ref{fig:groups} averages the four pair-level contrasts within each group and configuration. Tables~\ref{tab:groups} and~\ref{tab:zero} report group means and zero-shot results. Displayed precision is not an uncertainty estimate.

\noindent\begin{minipage}{\columnwidth}
\begin{center}
\captionsetup{hypcap=false}
\includegraphics[width=\columnwidth]{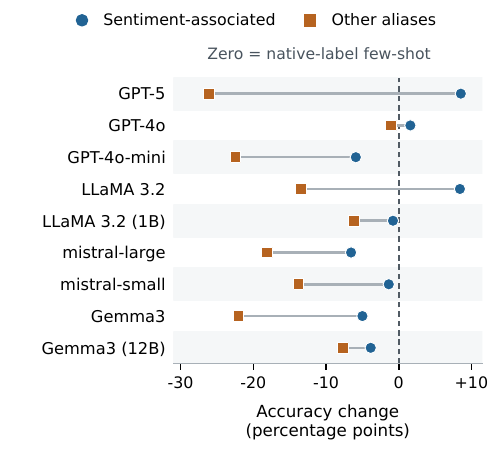}
\captionof{figure}{Group-mean accuracy changes from the attacked native-label few-shot baseline. Each marker averages four pair-level differences; the line connects the groups for one configuration. Zero means equal accuracy to native labels, not zero accuracy. Sentiment-associated means exceed other-alias means in all nine rows and exceed the baseline in three. These are descriptive comparisons, not significance tests.}
\label{fig:groups}
\end{center}
\end{minipage}

\newpage
\noindent\begin{minipage}{\columnwidth}
\begin{center}
\captionsetup{hypcap=false}
\small
\setlength{\tabcolsep}{3.3pt}
\begin{tabular}{lrrr}
\toprule
Recorded configuration & N & S & O \\
\midrule
GPT-5 & 49.44 & \textbf{57.97} & 23.34 \\
GPT-4o & 88.44 & \textbf{90.03} & 87.39 \\
GPT-4o-mini & 89.00 & 83.09 & 66.55 \\
LLaMA 3.2 & 64.25 & \textbf{72.66} & 50.75 \\
LLaMA 3.2 (1B) & 50.81 & 50.02 & 44.63 \\
mistral-large & 84.56 & 78.00 & 66.45 \\
mistral-small & 86.81 & 85.44 & 73.02 \\
Gemma3 & 79.31 & 74.31 & 57.27 \\
Gemma3 (12B) & 85.19 & 81.33 & 77.55 \\
\bottomrule
\end{tabular}

\captionof{table}{Attacked few-shot mean accuracy (\%). N: native labels; S: mean of four sentiment-associated pairs; O: mean of four other pairs. Bold values exceed N in the same row. Subtracting N from S and O gives the markers in Figure~\ref{fig:groups}, up to display rounding.}
\label{tab:groups}
\end{center}
\end{minipage}

\noindent\begin{minipage}{\columnwidth}
\begin{center}
\captionsetup{hypcap=false}
\small
\setlength{\tabcolsep}{4pt}
\begin{tabular}{lrrr}
\toprule
Recorded configuration & Clean & Attacked & Drop \\
\midrule
GPT-5 & 93.00 & 47.50 & 45.50 \\
GPT-4o & 90.50 & 85.00 & 5.50 \\
GPT-4o-mini & 88.00 & 82.50 & 5.50 \\
LLaMA 3.2 & 89.00 & 67.50 & 21.50 \\
LLaMA 3.2 (1B) & 73.00 & 50.00 & 23.00 \\
mistral-large & 90.50 & 82.00 & 8.50 \\
mistral-small & 91.00 & 81.50 & 9.50 \\
Gemma3 & 87.00 & 69.50 & 17.50 \\
Gemma3 (12B) & 89.50 & 71.50 & 18.00 \\
\bottomrule
\end{tabular}

\captionof{table}{Native-label zero-shot accuracy (\%). Clean omits the attack suffix; Drop is clean minus attacked accuracy (pp). These prompts have no demonstrations and are separate from the few-shot reference used to compare aliases.}
\label{tab:zero}
\end{center}
\end{minipage}

\clearpage
\section{Protocol and Evidence Status}
\label{app:protocol}

\paragraph{Labels and intended mapping.}
The punctuation pair is printed as \texttt{@\#\$/\textasciicircum} versus \texttt{*\&\%!}. This transcribes the retained notation, not verified runtime bytes. The example \emph{green}$\mapsto$positive and \emph{red}$\mapsto$negative illustrates the intended transformation. Complete serialized mappings, including the handling of unexpected outputs, are unavailable. Figure~\ref{fig:overview} explains the intended workflow without reconstructing a historical prompt.

\paragraph{Data and model provenance.}
The protocol supplies rating criteria and selection order, but not review IDs or auditable split assignments. We cannot verify absence of demonstration--evaluation leakage. Table row names, including parenthetical size suffixes, are retained identifiers. For example, LLaMA~3.2 and LLaMA~3.2 (1B) are separate recorded rows; the name without a suffix does not establish a specific model size. These labels do not support claims about scaling or distillation.

\paragraph{Missing inference details.}
Unavailable fields include provider/checkpoint snapshot, execution date, decoding settings, seeds or repeat counts, output-length and truncation policies, message boundaries, complete prompts, and output parsing. Exact versus substring matching, fallback rules, refusals, and extra text can all affect alias accuracy. We do not substitute plausible defaults for historical facts.

\newpage
\noindent\begin{minipage}{\columnwidth}
\begin{center}
\captionsetup{hypcap=false}
\small
\begin{tabular}{lcc}
\toprule
Prompt condition & Clean & Attacked \\
\midrule
Native, zero-shot & Score & Score \\
Native, few-shot & --- & Mean \\
Alias, few-shot & --- & Mean \\
\bottomrule
\end{tabular}
\captionof{table}{Available accuracy summaries. A dash means unavailable in the retained record. A few-shot mean summarizes shot settings; aggregation across demonstration orders cannot be verified.}
\label{tab:coverage}
\end{center}
\end{minipage}

\paragraph{Available comparisons.}
Table~\ref{tab:coverage} distinguishes the available conditions. Clean zero-shot scores establish performance without either demonstrations or the suffix. They cannot replace clean few-shot scores when estimating how aliasing changes the loss caused by the attack, as Equation~\ref{eq:decomposition} requires.

\paragraph{Aggregation and reproducibility.}
The numerical table describes averages over four shot settings; the protocol also describes two orderings. Their weighting and condition-level variation cannot be audited. The native/alias comparison relies on the reported use of common few-shot conditions; implementation-level matching cannot be checked. The present analysis gives each of the four pair-level means equal weight within a group and subtracts the corresponding row's native-label baseline. This arithmetic does not reconstruct missing outcomes or reproduce the original model runs. Separate shot, order, and item-level results cannot be inferred from the means.

\end{document}